\pdfoutput=1
\documentclass[sigconf, nonacm]{acmart}

\begin{document}

\title{MIRAGE: A Micro-Interaction Relational Architecture for Grounded Exploration in Multi-Figure Artworks}

\author{Jui-Cheng Chiu}
\email{chiu119@purdue.edu}
\affiliation{%
  \institution{Purdue University}
  \city{West Lafayette}
  \state{Indiana}
  \country{USA}
}

\author{Yu-Chao Wang}
\email{wang3368@purdue.edu}
\affiliation{%
  \institution{Purdue University}
  \city{West Lafayette}
  \state{Indiana}
  \country{USA}
}

\author{Shengyang Luo}
\email{luo525@purdue.edu}
\affiliation{%
  \institution{Purdue University}
  \city{West Lafayette}
  \state{Indiana}
  \country{USA}
}

\author{Tongyan Wang}
\email{wang5298@purdue.edu}
\affiliation{%
  \institution{Purdue University}
  \city{West Lafayette}
  \state{Indiana}
  \country{USA}
}

\author{Qi Yang}
\email{yang2767@purdue.edu}
\affiliation{%
  \institution{Purdue University}
  \city{West Lafayette}
  \state{Indiana}
  \country{USA}
}

\author{Nabin Khanal}
\email{khanaln@purdue.edu}
\affiliation{%
  \institution{Purdue University}
  \city{West Lafayette}
  \state{Indiana}
  \country{USA}
}

\author{Yingjie Victor Chen}
\email{chen489@purdue.edu}
\affiliation{%
  \institution{Purdue University}
  \city{West Lafayette}
  \state{Indiana}
  \country{USA}
}

\renewcommand{\shortauthors}{Chiu et al.}

\begin{abstract}

Appreciating multi-figure paintings requires understanding how characters relate through subtle cues like gaze alignment, gesture, and spatial arrangement. We present \textit{MIRAGE}, an evidence-centric framework designed to scaffold the exploration of these "micro-interactions" in multi-figure artworks. While such cues are essential for deep narrative appreciation, they are often distributed across complex scenes and difficult for viewers to systematically identify. Existing vision-language models (VLMs) frequently fail to provide reliable assistance, offering ungrounded interpretations that lack traceable visual evidence.

MIRAGE addresses this by constructing a structured intermediate representation capturing identities, pose cues, and gaze hypotheses. However, the challenge extends beyond extracting these cues to coordinating them during interpretation. Without an explicit mechanism to organize and reconcile relational evidence, models often collapse multiple interaction hypotheses into a single unstable or weakly grounded narrative, even when low-level signals are available. This representation allows users to verify how high-level interpretations are anchored in low-level visual facts.

By separating spatial grounding from narrative generation, MIRAGE enables users to inspect and reason about figure-to-figure relationships through a verifiable evidence layer. We evaluate MIRAGE against painting-only VLM baselines using a blind assessment protocol. Results show that MIRAGE significantly improves identity consistency, reduces relational hallucinations, and increases the coverage of subtle interactions. These findings suggest that structured grounding can serve as a critical interaction control layer, providing the necessary scaffolding for a more reliable, transparent, and human-led understanding of complex visual narratives.

\end{abstract}



\begin{teaserfigure}
  \includegraphics[width=\textwidth]{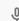}
  \caption{MIRAGE: Grounded interaction through structured visual representation. (Left) The system transforms raw visual signals into inspectable grounding layer via VLM-guided refinement, encoding relations, roles, and geometric evidence. (Right) This part representation enables a collaborative interaction where users engage in multi-step dialogue, challenge the system's interpretations, and receive transparent, evidence-linked responses. The example is based on \textit{Katrina} by Jonathan Hodge.}
  \Description{Teaser figure for our pipeline}
  \label{fig:teaser}
\end{teaserfigure}

\maketitle

\section{Introduction}

Art appreciation involves complex perceptual and cognitive processes shaped by visual organization \cite{leder2004model}. In multi-figure paintings, interpreting the intended narrative often depends on perceiving \textit{micro-interactions}: fine-grained relational cues such as gaze alignment, gesture contact, body orientation, and spatial proximity that suggest social intent, emotional tension, and narrative flow. Prior work demonstrates that humans infer social interactions from such visual inputs \cite{McMahon2023, williams2024integration, gaze_social_signal}, and that these local relational details shape how viewers understand interpersonal dynamics in art \cite{berger1972ways, freedberg1989power}. However, in complex compositions, these signals are distributed across many figures and regions of the canvas. Consequently, viewers may struggle to systematically identify these cues, making it challenging to move from a broad impression to a grounded reading of how specific figures relate to one another.

Recent advancements in vision-language models (VLMs) and AI-assisted art systems demonstrate strong capabilities in holistic artwork description and conversational exploration \cite{yuan2024artgpt, khadangi2025cognartive, li2024deep, shende2024aienhanced, lee2024llava, zheng2025artmentor}. However, these models are rarely architecturally designed to maintain the fine-grained relational grounding necessary for micro-interaction analysis. The challenge further extends beyond detecting such cues to coordinating them during interpretation. Without an explicit mechanism to organize and reconcile relational evidence, models often collapse multiple interaction hypotheses into unstable or weakly grounded interpretations, even when the underlying signals are correctly identified. When interpreting overlapping poses or multiple candidate interactions, VLM outputs frequently drift toward vague narrative summaries, unstable identity references, and merged attributes \cite{ferrato2025exploring, rachabatuni2024context}. Because these systems generally operate as opaque models that bypass structured spatial analysis, they often fail to provide the traceable visual evidence needed to verify or deepen an interpretive process \cite{arita2025assessing, wang2025pinning}. This instability limits their utility for user-centered exploration, particularly for laymen museum visitors and learners who rely on understanding who is acting, who is responding, and what visual evidence supports that reading.

To address this gap, we present \textit{MIRAGE}, a \textit{Micro-Interaction Relational Architecture for Grounded Exploration in Multi-Figure Artworks}. Positioned as an evidence-centric framework, MIRAGE is motivated by the premise that the interpretation of complex human scenes benefits from stabilizing character identities, spatial relations, and local interaction cues before high-level reasoning occurs. MIRAGE constructs a structured intermediate representation for further interpretations. Rather than improving perception alone, MIRAGE introduces an explicit control layer that stabilizes how relational evidence is selected, combined, and interpreted. This representation captures object anchors, pairwise relations, and localized evidence, effectively separating geometric grounding from semantic interpretation. By establishing these relations as a foundational layer, MIRAGE allows generated appreciation guidance to remain tied to identifiable visual evidence rather than relying solely on global scene impressions.

As an interface framework, MIRAGE facilitates interpretive collaboration by exposing subtle figure-to-figure dynamics as inspect\-able evidence. We evaluate MIRAGE by comparing its outputs against direct image-only prompting of a general-purpose VLM, focusing on the recovery and interpretability of human-recognizable micro-interactions. The contributions of this work are:

\begin{enumerate}
    \item an evidence-centric architecture that decouples grounding from downstream interpretation, enabling controlled reasoning over ambiguous multi-figure interactions;
    \item a relationally grounded representation that provides users with verifiable visual evidence to support the exploration of subtle character dynamics; and
    \item a systematic evaluation demonstrating that the extra layer of grounding yields more specific, traceable, and micro-interaction-aware analysis compared to direct image-only VLM prompting.
\end{enumerate}

\section{Related Work}

This research intersects with art appreciation, VLMs, and grounded scene understanding. We review prior work along five directions: (1) art appreciation as a relational and interpretive process, (2) LLM and VLM systems for artwork understanding, (3) grounding and structured intermediate representations for multimodal reasoning, (4) micro-interaction in multi-figure artworks, and (5) interactive dialogue for collaborative art exploration. Together, these threads underscore the value of a grounding-centered approach to support the fine-grained interpretation of complex multi-figure scenes.

\subsection{Art Appreciation as Relational Interpretation}

Art appreciation involves interpreting how visual elements form meaningful relationships within a composition. Prior work shows that artworks are perceived as structured configurations rather than isolated parts, with meaning emerging through form, spatial arrangement, and perceptual emphasis \cite{arnheim1974art}. Aesthetic experience further relies on stages of interpretation and evaluation that build upon immediate perception \cite{leder2004model}. In multi-figure paintings, this process is particularly relational, as viewers infer narrative and social meaning through cues such as gaze, gesture, orientation, and proximity. Prior work on social scene perception indicates that observers tend to extract interactional structure from visual information \cite{McMahon2023}, integrating head and body cues to perceive dyadic interaction \cite{williams2024integration}. Moreover, attention is frequently drawn toward socially interactive regions in naturalistic scenes \cite{skripkauskaite2023attentional}. Consequently, prior work suggests that understanding multi-figure artworks depends on perceiving fine-grained visual relations rather than merely identifying depicted subjects.

\subsection{LLMs and VLMs for Artwork Interpretation}

Recent advances in LLMs and VLMs facilitate novel forms of computational support for artwork interpretation, including educational guidance, formal analysis, question answering, and evaluative dialogue. For example, \textit{LLaVA-Docent} demonstrates how a multimodal LLM can support art appreciation education through personalized conversational tutoring \cite{lee2024llava}. Concurrently, \textit{GalleryGPT} and \textit{ArtGPT-4} indicate that art-oriented multimodal models may produce richer formal analyses and enhance performance on artwork understanding tasks \cite{bin2024gallerygpt,yuan2024artgpt}. Related work exploring artwork question answering and context-aware cultural heritage chatbots suggests that integrating external knowledge and contextual descriptions improves responses compared to purely image-based approaches \cite{ferrato2025exploring,rachabatuni2024context}. Furthermore, systems such as \textit{ArtMentor} and recent studies on critique generation suggest that LLMs are increasingly capable of participating in artwork evaluation and interpretive dialogue \cite{zheng2025artmentor,arita2025assessing}. However, existing models frequently exhibit limitations when interpretation relies on contextual accuracy, fine-grained evidence, or socially embedded reasoning. They commonly encounter challenges with visual ambiguity, symbolic meaning, and maintaining consistency across related queries \cite{ferrato2025exploring,arita2025assessing,pado2025artwork}. MIRAGE extends prior advances in AI-assisted art interpretation by addressing these limitations through a grounding-centered architecture, which systematically separates relational evidence construction from downstream semantic generation.

\subsection{Grounding and Structured Intermediate Representations}

Recent research in vision-language reasoning indicates that explicit intermediate representations may improve reliability and interpretability by separating perceptual extraction from downstream inference. In computer vision, scene graph approaches represent images as entities and relations, providing a stable substrate for relational reasoning in complex scenes. Foundational works such as \textit{Image Generation from Scene Graphs}~\cite{johnson2018imagegenerationscenegraphs}, \textit{Neural Motifs}~\cite{zellers2018neuralmotifsscenegraph}, and \textit{GQA}~\cite{hudson2019gqanewdatasetrealworld} illustrate how entity-centric representations support consistent and compositional reasoning.

In parallel, multimodal systems increasingly adopt modular and interpretable pipelines over purely end-to-end generation. Modern approaches such as \textit{Visual ChatGPT}~\cite{wu2023visualchatgpttalkingdrawing} and \textit{ViperGPT}~\cite{suris2023vipergptvisualinferencepython} introduce intermediate stages that decompose perception and reasoning, aiming to enhance transparency and controllability compared to opaque one-shot inference. Similarly, \textit{Chain-of-Thought} prompting underscores the value of exposing intermediate reasoning steps in language models~\cite{wei2023chainofthoughtpromptingelicitsreasoning}. From an HCI perspective, this shift aligns with the growing emphasis on  inspectable AI systems. \textit{Guidelines for Human-AI Interaction}~\cite{amershi2019guidelines} suggest that users benefit from understanding system behavior and assessing output reliability, motivating designs that surface structured evidence rather than only final results. However, prior work largely targets functional tasks such as visual question answering or navigation, frequently overlooking the subtle relational cues necessary for art interpretation. In contrast, visual meaning in artworks heavily relies on compositional relationships such as gaze, spatial grouping, and tension. Arnheim’s \textit{Art and Visual Perception}~\cite{arnheim1974art} discusses how such relational structures shape human interpretation.

Building on these insights, MIRAGE introduces a \textit{Grounding Document}, an inspectable intermediate representation that integrates character identities, spatial anchors, and localized relational evidence prior to semantic interpretation. By formulating grounding as a queryable layer, MIRAGE bridges structured scene representation, modular multimodal reasoning, and human-centered interpretability to facilitate the exploration of complex multi-figure artworks.

\subsection{Micro-Interaction in Multi-Figure Artworks}

We define \textit{micro-interactions} as localized, fine-grained relational configurations between figures, or between figures and salient objects, that shape scene interpretation. These interactions emerge from subtle visual cues---such as gaze, gesture, posture, and proximity---that are extensively studied as \textit{social signals} in human perception~\cite{poggi2012socialsignals}. Because observers naturally integrate multiple weak signals to infer interpersonal dynamics~\cite{ambady1992thin}, relational meaning rarely stems from a single isolated cue. For example, while gaze is critical for coordinating shared attention~\cite{hessels2020gaze, stephenson2021gaze}, it must be contextually combined with other spatial signals to convey complex narratives like support or tension.

Despite their importance, current computational approaches to art understanding heavily prioritize global captioning. By overlooking localized relational structures, fine-grained interactions are frequently omitted, inconsistently described, or hallucinated during downstream interpretation. MIRAGE addresses this limitation by formalizing micro-interactions as structured, inspectable units grounded in verifiable visual evidence. Through the explicit representation of stable character identities, spatial anchors, and cue-level signals, the framework provides the necessary scaffolding for users to systematically reason about subtle relationships that holistic descriptions otherwise obscure.

\subsection{Interactive Dialogue for Art Exploration}

Conversational systems support interactive sensemaking by enabling users to iteratively probe system outputs, explore multiple perspectives, and incrementally construct meaning~\cite{amershi2019guidelines, park2023generativeagentsinteractivesimulacra, pirolli2005sensemaking}. This dynamic exploration is particularly crucial for art interpretation, where understanding emerges over time rather than through single-pass observation. However, many existing systems rely on free-form generation without explicit visual grounding. Because effective human-AI interaction requires transparency and reliability to build trust~\cite{amershi2019guidelines}, ungrounded conversational agents often fail to ensure interpretations remain consistent, verifiable, and strictly tied to specific visual details in complex multi-figure scenes.

MIRAGE extends this prior work by anchoring conversational dialogue within a queryable evidence layer that incorporates stable character identities, object anchors, and relation-level cues. Rather than treating dialogue as a strictly generative process, this approach assists users in iteratively exploring artworks while maintaining an inspectable connection between high-level semantic interpretations and localized visual evidence.

\section{System Architecture}

MIRAGE supports micro-interaction understanding in multi-figure paintings by introducing a structured grounding representation prior to interpretation. Instead of relying on end-to-end captioning, MIRAGE decomposes a painting into character anchors with stable identities, objects that organize interaction, geometric interaction cues, and pairwise relation states, which are consolidated into a grounding document for downstream reasoning. This design enables interpretation over explicit, inspectable interaction structure rather than implicit inference from the raw image.

\subsection{Face-First Character Detection \& Object Anchoring}

MIRAGE initializes a structured grounding space by identifying both character and object anchors. Rather than relying on body-first detection, we adopt a face-first strategy in which faces serve as stable identity seeds and are later expanded to body-level regions. This design prioritizes identity stability in paintings, where full-body cues are often unreliable due to stylization, occlusion, and non-photographic poses.

Faces are detected using \textit{YOLOv8-face} \cite{yaseen2024yolo}, generating face boxes as follows, where $(x_i, y_i)$ is the upper-left corner, $(w_i, h_i)$ denote width and height, and $s_i$ is the confidence score:
\begin{equation}
F=\{f_i\}_{i=1}^{N_f}, \; f_i=(x_i, y_i, w_i, h_i, s_i) \,.
\end{equation}
Confidence filtering, non-maximum suppression, and lightweight geometric constraints are applied to remove implausible detections caused by painterly texture or fragmented visual structure. The resulting set $F$ provides a set of identity-stable anchor points that index distinct characters in the scene.

To recover character support regions, MIRAGE detects body boxes and associates each face with the most plausible region:
\begin{equation}
B=\{b_j\}_{j=1}^{N_b} \,.
\end{equation}
Let $c(f_i)$ denote the center of face $f_i$. The selected body box is defined as
\begin{equation}
b^*(f_i)=\arg\max_{b_j \in B} S(f_i,b_j) \,,
\end{equation}
which is subject to the constraint that $c(f_i)$ lies inside the body box and within its upper half. This enforces a head-location prior within the person region. When multiple candidates exist, selection is based on detector confidence and spatial consistency. If no valid match is found, the face region is expanded into a fallback body estimate to ensure a usable character crop.

This step produces a set of character-aligned regions $b^*(f_i)$ that serve as spatial supports for each identity anchor. These regions are later used to extract pose, appearance, and interaction cues while maintaining alignment between identity and geometry.

In parallel, MIRAGE extracts object anchors to capture interactive scene structures. A generic detector first proposes candidate objects, which are then filtered by GPT-5.4 to retain semantically central elements such as tombs, inscriptions, and staffs. Unlike standard detection, which favors common categories, this step prioritizes objects that organize interaction, including elements that figures point to, touch, or attend to.

The resulting object set provides a set of interaction-relevant reference points that complement character anchors. Together, face-based character anchoring and object grounding form the grounding initialization, defining a unified grounding space where identities and interaction-relevant entities are explicitly represented. This initialization establishes the basis for subsequent character-centric grounding and relation modeling.

\subsection{Character-Centric Grounding Construction}

After grounding initialization, MIRAGE converts each detected figure into a structured character profile,
\begin{equation}
p_i = (\mathrm{id}_i, b_i, f_i, \ell_i, a_i, n_i) \,,
\end{equation}
where $\mathrm{id}_i$ is a stable character ID, $b_i$ and $f_i$ denote body and face regions, $\ell_i$ encodes spatial attributes, $a_i$ captures appearance and posture cues, and $n_i$ represents local neighborhood context. This representation produces identity-indexed character states that preserve alignment between local evidence and global scene structure.

MIRAGE augments each profile with geometric interaction cues using \textit{YOLOv8x-pose}. Instead of raw keypoints, pose outputs are abstracted into higher-level signals such as posture, orientation, and candidate pointing direction. Given shoulder and wrist points $s_i, w_i$, a pointing ray is defined as
\begin{equation}
r_i(t)=w_i+t(w_i-s_i), \; t\ge 0 \,,
\end{equation}
and the first intersected character or object is recorded as a candidate interaction target. These cues provide quantifiable but ambiguous geometric hypotheses about attention and interaction.

\begin{figure}[t]
    \centering
    \includegraphics[width=0.445\linewidth]{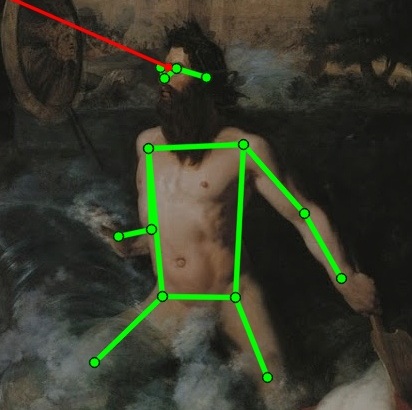}
    \includegraphics[width=0.42\linewidth]{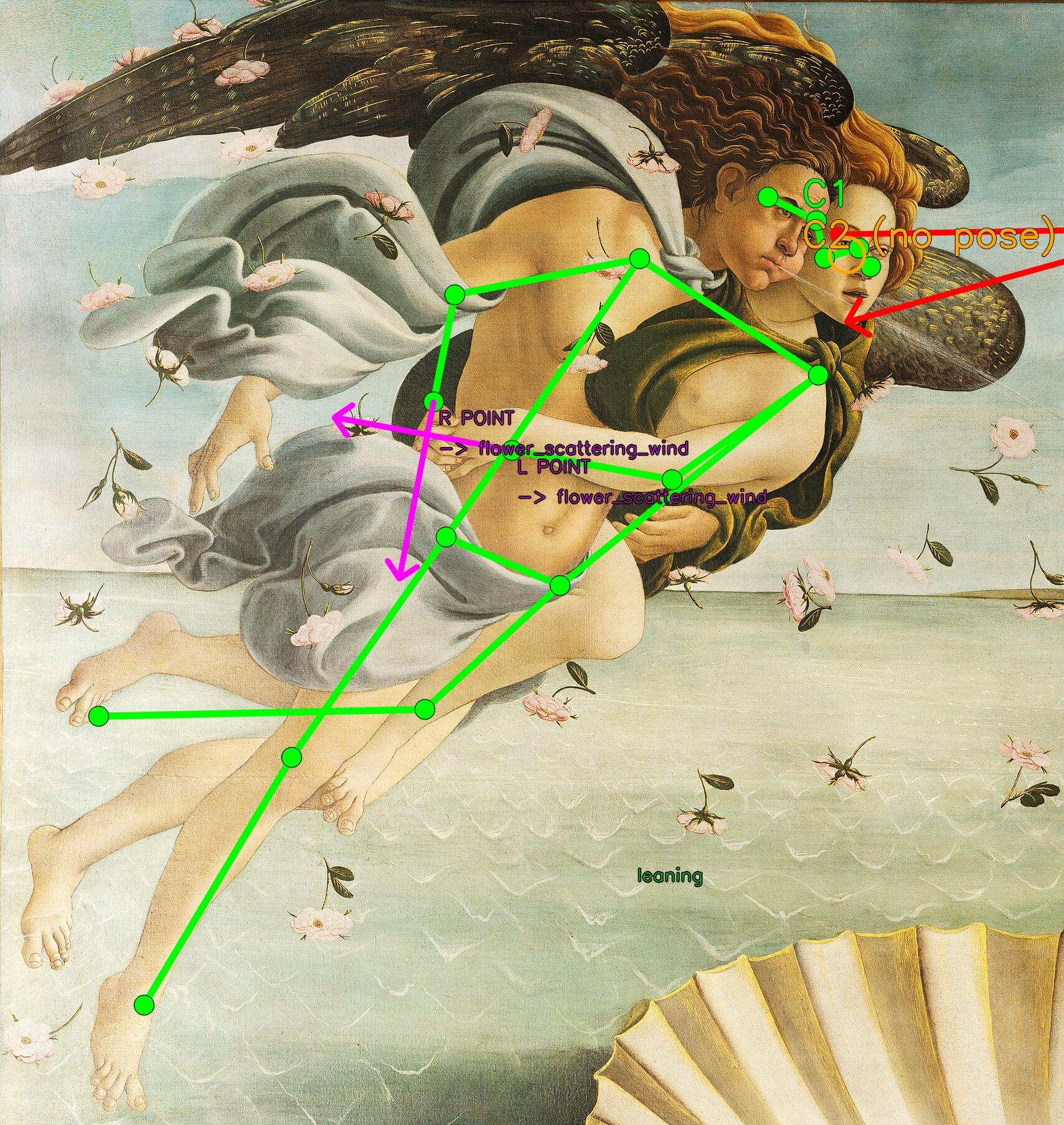}
    \caption{Pose and gaze estimation conditioned on grounding reliability. Left: when both face and body regions are available, MIRAGE recovers full skeletal structure and derives gaze cues. Right: when body evidence is incomplete (e.g., C2), pose inference is selectively disabled to prevent unreliable geometric interpretation. Examples are drawn from \textit{Achille poursuivi par le Xanthe} (Frédéric Henri Schopin) and \textit{The Birth of Venus} (Sandro Botticelli)}
    \Description{Examples of pose and gaze estimation based on regional availability.}
    \label{fig:character_pose_examples}
\end{figure}

The reliability of these geometric cues depends on grounding availability. As shown in Figure~\ref{fig:character_pose_examples}, when both face and body regions are detected, MIRAGE recovers a full skeleton and produces usable pose and gaze estimates. When body support is missing, pose inference is disabled to avoid introducing unsupported geometric signals. This conditional use of geometry ensures that only grounded structural evidence is propagated to later stages.

To resolve ambiguity in geometric hypotheses, MIRAGE performs VLM-based semantic enrichment using GPT-5.4. Each character is described under a stable ID using both local crops and the full image, focusing on interaction-relevant attributes like body language, facial expression, object interaction, and attention direction. Geometry provides candidate interaction structure, while the VLM refines and disambiguates these signals under visual context.

Finally, MIRAGE extracts scene-level context (e.g., composition and setting) to support the interpretation of local interactions. Together, character profiles, geometric cues, and VLM enrichment produce interaction-ready character states that serve as the basis for subsequent relational grounding.

\subsection{Relational Grounding}

Given interaction-ready character states, MIRAGE constructs pairwise relation records for all character pairs. Instead of directly producing free-form descriptions, relations are first represented as structured evidence derived from spatial and geometric cues.

For each pair $(i,j)$, MIRAGE records normalized inter-character distance, bounding-box overlap (IoU), relative position, and size ratio. These quantitative metrics are combined with candidate interaction cues derived from pose direction vectors, proximity thresholds, and shared object references. A localized pair crop, defined by the union of the two character bounding boxes with a 15\% spatial margin, is extracted to preserve relation-specific visual context.

\begin{figure}[t]
    \centering
    \includegraphics[width=\linewidth]{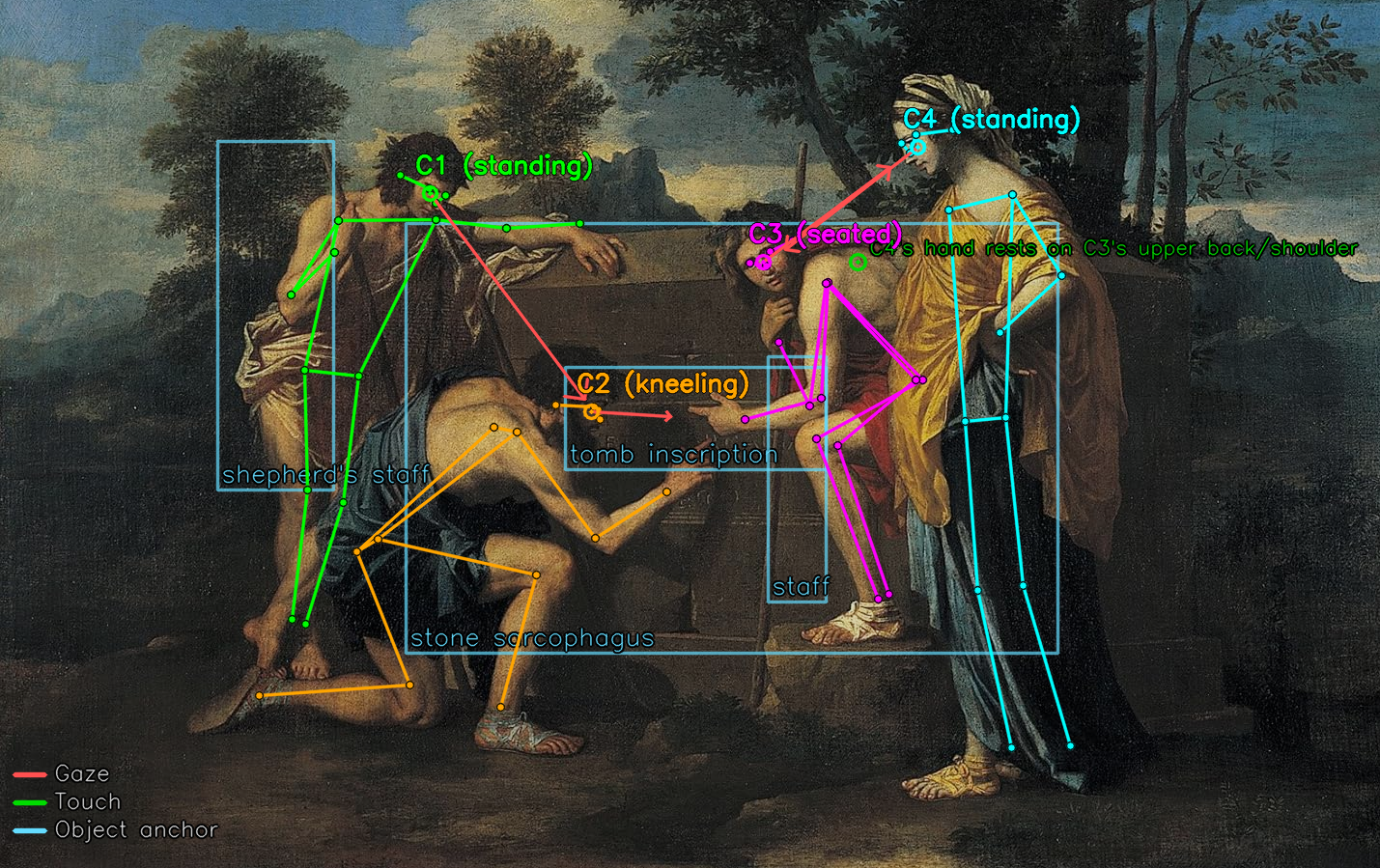}
    \caption{Relational grounding in MIRAGE. Pairwise relations are constructed from character-centric signals, including spatial overlap, gaze direction, and proximity. The visualization makes explicit how interaction-relevant structure (e.g., contact, attention, and object-centered relations) is derived prior to grounding document assembly, enabling inspection of how interactions are grounded in localized visual evidence. Example from \textit{Et in Arcadia Ego} (Nicolas Poussin).}
    \Description{Diagram illustrating the construction of pairwise relations from character signals.}
    \label{fig:stage3_relations}
\end{figure}

This process yields relation states encoding interaction-aware structures. Unlike single-character representations, these states capture the foundational micro-interactions emerging from relative configurations. By storing each relation as a structured record, MIRAGE establishes a relational reference for downstream language generation, serving as a set of algorithmically derived anchor points. Unlike conventional conversational agents that rely solely on unconstrained generation \cite{wang2025pinning, zheng2025artmentor}, this structural representation mitigates the premature collapse of foundational evidence into vague natural-language summaries, ensuring that the subsequent generative process builds upon localized, inspectable signals.

\subsection{Grounding Document Assembly}

MIRAGE consolidates the extracted scene-level, character-level, and relation-level evidence into a unified grounding document formatted in Markdown. It serves as a structured intermediate representation, organizing heterogeneous pipeline signals into a single, coherent reference baseline for downstream reasoning.

The Markdown schema integrates three primary domains into a continuous textual structure. Scene Context encodes global composition and semantic tags. Character Profiles store stable identifiers (e.g., C1, C2), spatial attributes, and pose-derived signals. Concurrently, Pairwise Relations capture localized interaction evidence alongside shared object anchors.

The selection of Markdown as the structural interface is based on recent empirical findings regarding LLM performance. Studies indicate that macro-formatting significantly influences LLM comprehension, with well-structured formats like Markdown improving reasoning accuracy in complex text processing tasks \cite{he_does_2024, braun_hidden_2025}. Furthermore, LLMs exhibit notable sensitivity to spurious features and arbitrary formatting variations \cite{sclar_quantifying_2024}. By standardizing heterogeneous visual signals into a predictable Markdown schema, MIRAGE mitigates formatting-induced performance fluctuations, making geometric data more accessible and stable for VLM processing.

This grounding document acts as the primary interface between visual analysis and generative interpretation. By treating algorithmic signals as a foundational reference, the document allows the VLM to anchor its open-ended semantic interpretations and artistic reasoning to geometric constraints. This structured approach reduces the likelihood of errors associated with inferring complex, multi-figure geometries directly from raw pixels.

\paragraph{Geometry–VLM Integration and Conflict Preservation}
MIRAGE uses a geometry-first design with VLM refinement. Geometry provides spatial structure, while VLMs resolve higher-level cues such as gaze and intent that are ambiguous from 2D information alone. When disagree, MIRAGE preserves both interpretations instead of forcing a single decision. This allows the system to expose uncertainty and support reasoning over multiple possible interactions.

\subsection{Grounded VLM Interpretation}

In the final architectural stage, MIRAGE employs a Vision-Language Model (GPT-5.4) as its grounded interpretation engine. Building upon the structural interface established in the previous stage, the generation process shifts away from analyzing raw pixels in isolation---a method inherently prone to identity blending and spatial hallucination in stylized, multi-figure compositions. Instead, the VLM executes its interpretive reasoning directly over the assembled Markdown grounding document. This document defines a structured reasoning space in which identities, relations, and evidential constraints are explicitly encoded, limiting the model's inference to grounded interaction hypotheses.

This generative phase is guided by targeted system prompts that enforce constrained extrapolation over the structured grounding representation. The model is instructed to treat the provided schema as an immutable reference baseline. Operating strictly within these algorithmic boundaries, the VLM engages in abductive reasoning to articulate implicit emotional dynamics, interpret subtle visual nuances omitted by pose estimators, and synthesize broader art historical contexts. This establishes an explicit control layer over how relational evidence is selected, combined, and interpreted during generation. The primary constraint requires that higher-order narrative inferences logically align with the derived spatial and semantic anchors, effectively preventing the invention of unsupported relational dynamics.

When processing the intentional modality conflicts preserved within the document, the system prompts the VLM to evaluate these algorithmic disagreements as explicit indicators of visual ambiguity. Rather than attempting a deterministic override, the model conceptually integrates these discrepancies, recognizing them as potential artifacts of complex occlusions, stylized drapery, or deliberate compositional tension inherent to the artwork.

Ultimately, this reasoning mechanism yields nuanced, evidence-based scene narratives and pairwise interaction analyses. While this scaffolded approach significantly enhances structural fidelity, the system's overall reliability naturally remains bounded by the performance characteristics of the underlying detection and pose estimation pipelines.

\section{System Capabilities}

\begin{figure*}[t]
  \centering
  \includegraphics[width=\textwidth]{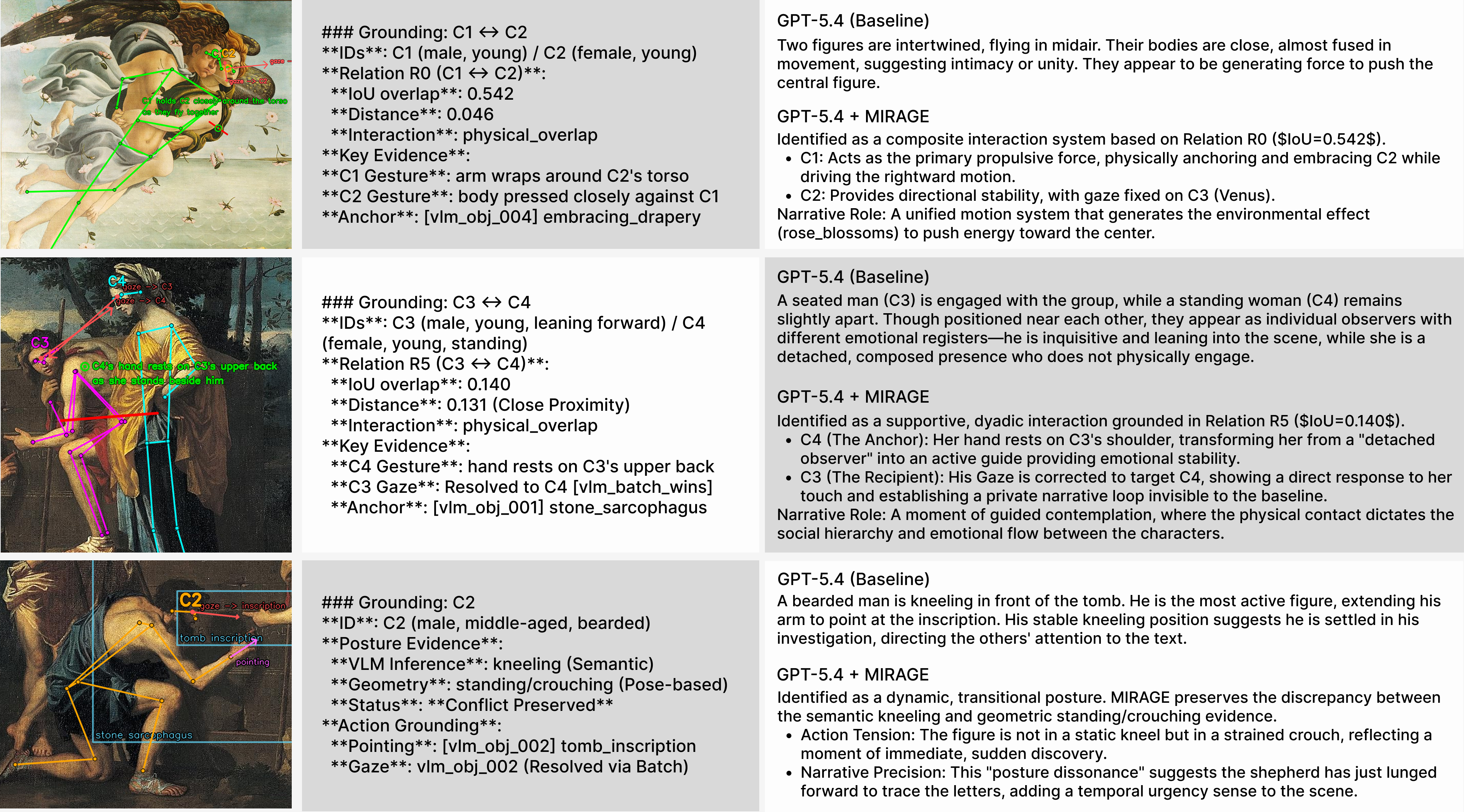}
  \caption{Comparison between baseline GPT-5.4 and MIRAGE-enhanced interpretation. Across representative interaction scenarios, MIRAGE produces identity-consistent, directional, and evidence-supported interpretations, while baseline outputs tend to be generic, ambiguous, or lack grounding. \textbf{Left}: Visual grounding with pose and gaze vectors. \textbf{Middle}: Structured MD document snippets highlighting key metrics such as $IoU$ and posture discrepancies. \textbf{Right}: Interpretation differences from qualitative observation to structured, evidence-based analysis across three scenarios: character de-confliction, micro-interaction recovery, and conflict preservation. Example from \textit{The Birth of Venus} (Sandro Botticelli) and \textit{Et in Arcadia Ego} (Nicolas Poussin).}
  \Description{A qualitative comparison figure showing three columns. The left column displays visual grounding with pose and gaze vectors overlaid on multi-figure artwork crops. The middle column shows structured Markdown document snippets highlighting metrics such as IoU and posture discrepancies. The right column details interpretation differences, contrasting the baseline GPT-5.4 model with the MIRAGE-enhanced model across three scenarios: character de-confliction, micro-interaction recovery, and conflict preservation.}
  \label{fig:main_results}
\end{figure*}

MIRAGE provides a structured representation of multi-character scenes by systematically organizing visual evidence---such as character identity, pose, gaze, spatial relations, and object-mediated interactions---into distinct relational components. The system builds an interpretable framework for interaction analysis. To illustrate these capabilities, we present a qualitative comparison between a baseline VLM and the MIRAGE-augmented model across several representative artwork scenarios, detailed in Figure \ref{fig:main_results}. The following sections explore how this structured grounding enhances interaction understanding across four conceptual levels: recovering fine-grained micro-interactions at the local level, modeling object-mediated connections at the intermediate level, preserving multimodal ambiguity at the representation level, and surfacing directional organization at the global level.

\subsection{Local: Fine-Grained Relations}

Standard VLMs often produce global summaries that lack structural representation of interactions between specific characters. MIRAGE addresses this limitation by constructing a grounding layer that encodes localized interaction cues derived from spatial and relational evidence, enabling stable and verifiable reasoning.

\paragraph{Deconstructing Composite Agents in High Overlap} In \textit{The Birth of Venus}, the airborne figures (C1, C2) exhibit substantial spatial overlap ($\mathrm{IoU}=0.542$), posing a challenge for maintaining distinct character identities. While a baseline GPT-5.4 provides a metaphorical interpretation describing them as ``fused in movement'', the MIRAGE-augmented model instead represents them as two distinct entities with strong spatial coupling. By leveraging this quantitative $\mathrm{IoU}$ data, the model captures their coordinated configuration and maintains a structured representation of their interaction.

\paragraph{Subtle Contact Recognition} In \textit{Et in Arcadia ego} (Fig. \ref{fig:main_results}, Row 2), MIRAGE identifies a localized interaction where C4 places a hand on C3’s back, supported by spatial overlap ($\mathrm{IoU}=0.140$). This is encoded as a discrete relation grounded in physical contact.

\subsection{Intermediate: Object Anchoring}

Interactions in multi-figure artworks are often organized around shared objects and focal points. MIRAGE models these structures by structurally linking characters through object anchors, which enables the representation of object-mediated interactions rather than treating characters as isolated entities.

\paragraph{Triangulated Attention via Anchors} In \textit{Et in Arcadia ego}, MIRAGE constructs a relation structure where C2 and C3's ``pointing'' gestures are both linked to the \textit{tomb\_inscription} in the center of the image (vlm\_obj\_002). By identifying this shared reference, the system treats the tomb as a central interaction anchor, organizing multiple character actions into a coherent relational topology rather than evaluating them as independent poses.

\paragraph{Object-Mediated Interaction} In \textit{The Birth of Venus}, the interaction between the two characters on the left is mediated by the \textit{embracing\_drapery} (vlm\_obj\_004). MIRAGE encodes this as a shared object-linked relation, where the artifact provides a structural connection between the figures, allowing the model to distinguish intentional object-mediated interaction from incidental spatial proximity.

\subsection{Representation: Conflict Awareness}

Rather than collapsing visual ambiguity into a single deterministic label, MIRAGE is designed to preserve conflicting signals that arise between the geometric pose estimation (YOLOv8x-pose) and the semantic enrichment (GPT-5.4) stages. This conflict preservation enables the system to retain multiple interpretations derived from different modalities, supporting transparent and evidence-aware reasoning.

\paragraph{Semantic vs. Geometric Dissonance} In \textit{Et in Arcadia ego} (Fig. \ref{fig:main_results}, Row 3), the VLM describes C2 as ``kneeling'' (semantic), whereas the pose estimation module infers a ``standing/crouching'' posture based on geometric cues. This discrepancy illustrates the potential divergence between high-level semantic interpretation and low-level spatial estimation.

\paragraph{Uncertainty-Aware Representation} Instead of forcing consensus, MIRAGE retains both interpretations under a \textit{Conflict Preserved} status. This approach allows downstream reasoning to operate over multiple hypotheses rather than relying on a single resolved label. As shown in the comparative evaluation, the baseline assumes a ``stable kneeling position,'' while the MIRAGE-enabled model incorporates the posture discrepancy into its interpretation, reflecting the inherent ambiguity in the observed pose rather than enforcing a fixed state.

\subsection{Global: Scene Topology}

Beyond isolated pairwise relations, MIRAGE captures higher-level interaction structures across the scene. By combining local relational signals, the system forms a global interaction topology that reflects how characters are spatially and directionally organized.

\paragraph{Directional Interaction Structure} The system identifies asymmetric configurations where multiple relational cues (e.g., gaze direction, pose orientation, spatial arrangement) converge toward a central figure, such as C3 in \textit{The Birth of Venus}. This convergence forms a structured pattern of directional relations, indicating that characters participate in scene-level organization rather than merely connecting locally.

\paragraph{Role-Like Structural Differentiation} The grounded representation enables the differentiation of characters based on their relational patterns. For example, some characters primarily contribute outgoing directional cues (e.g., gaze or motion toward others), while others serve as focal recipients or participate through object-mediated interactions (e.g., C4 in the Venus scene). This differentiation emerges organically from the relational structure rather than relying on pre-defined categorical labels.

Across the presented examples, MIRAGE shifts interaction understanding from implicit description to structured, evidence-based reasoning. At the local level, it recovers fine-grained micro-interactions utilizing localized spatial relations. At the intermediate level, it links characters through shared anchors, enabling the representation of object-mediated interactions. At the representation level, it preserves ambiguity across modalities, supporting conflict-aware reasoning without forcing premature resolution. At the global level, it organizes relational signals into coherent interaction structures. These capabilities address key challenges in multi-figure interaction understanding, particularly identity stability, relational ambiguity, and the representation of grounded interaction structures.

\section{Evaluation}

We evaluate MIRAGE through a two-part protocol examining output quality and user interaction. Section 5.1 quantitatively compares MIRAGE against image-only VLM baselines regarding identity consistency, relational accuracy, and micro-interaction coverage. Section 5.2 details a user study on how MIRAGE fosters evidence-based reasoning, transparency, and human-AI collaboration.

\subsection{Quantitative Evaluation}

We design our evaluation to directly measure the core challenges identified in multi-figure interaction understanding. Specifically, we assess (1) identity stability to address character-level drift, (2) interaction accuracy to evaluate the detection of meaningful relations, (3) directional reasoning to capture interaction structure and flow, and (4) grounding-aware reasoning to ensure interpretations are supported by explicit visual evidence.

\subsubsection{Dataset \& Protocol}

We evaluate whether structured grounding improves micro-interaction understanding in multi-figure artworks compared to non-grounded baselines. While VLMs often identify isolated perceptual cues, understanding micro-interactions requires integrating relational signals (e.g., gaze, posture, spatial configuration) into coherent interpretations. We designed a controlled evaluation protocol to measure identity consistency, interaction correctness, directional reasoning, and evidence usage.

Our dataset comprises 15 multi-figure images spanning three categories to ensure structural variety: 5 canonical paintings, 5 lesser-known artworks (to mitigate memorization bias), and 5 generated scenes. Each image contains 3 to 5 salient characters. We selected 5 to 7 key interaction instances per image, encompassing both pairwise and object-mediated relations. This yielded $N$ evaluation items, each paired with structured questions to systematically probe relational reasoning capabilities rather than relying on unconstrained descriptions.

To establish a ground truth, we constructed a reference set based on visually observable relational evidence rather than subjective narrative interpretation. For each interaction instance between characters $C_i$ and $C_j$, annotators labeled three dimensions: interaction existence (yes, no, or ambiguous), interaction type (categorized into visually grounded states like support, guidance, or shared attention), and directionality (initiator-to-receiver, mutual, or none). Annotations were strictly bound to observable cues such as gaze alignment, physical contact, or spatial proximity. Disagreements among annotators were resolved through consensus discussion.

\subsubsection{Evaluated Models}

We compare MIRAGE against a standard baseline using the same underlying model, GPT-5.4. The baseline condition utilizes GPT-5.4 with image-only input (the original painting), representing the standard VLM interaction paradigm. The MIRAGE condition provides GPT-5.4 with the original painting, the annotated image and the generated grounding document. For MIRAGE, the grounding document encodes stable character identities, resolved gaze targets, pairwise spatial relations, and object-level anchors, which serve as structured input to guide reasoning. Both conditions are evaluated under identical question templates and inference settings to ensure a fair comparison.

\subsubsection{Prompt Design for Evaluation}

For each interaction instance, the models are asked a series of structured questions designed to systematically probe relational reasoning. To evaluate interaction existence, the models are asked whether there is a meaningful interaction between $C_i$ and $C_j$. For the interaction type, they are prompted to identify the most likely interaction between the two characters based on visual evidence. To assess directionality, the models must specify who initiates the interaction, if anyone. Finally, to evaluate evidence grounding, the models are asked to describe the visual evidence that supports their interpretation. This design forces models to integrate multiple cues and produce directional, evidence-supported interpretations, rather than relying on isolated perception or generic descriptions.

\subsubsection{Results \& Analysis}

\begin{table}[t]
\centering
\small
\caption{Quantitative results on micro-interaction understanding across identity, interaction, direction, and grounding metrics. We compare our MIRAGE engine against the baseline model across four key metrics.}
\label{tab:results}
\setlength{\tabcolsep}{4pt}
\begin{tabular}{lcccc}
\toprule
Method & Identity & Interaction & Direction & Grounding \\
\midrule
GPT-5.4 (Base model) & 0.72 & 0.81 & 0.83 & 0.71 \\
MIRAGE (Ours) & \textbf{0.92} & \textbf{0.94} & \textbf{0.92} & \textbf{0.88} \\
\bottomrule
\end{tabular}
\end{table}

We evaluate performance using standard accuracy measures for interaction and direction, alongside task-specific metrics for identity stability and grounding-aware reasoning. Identity stability ($S_{\text{identity}}$) averages instance scores $s_i$ (1 for consistent references, 0.5 for minor ambiguity, 0 for identity drift) across $N$ evaluation items:
\begin{equation}
S_{\text{identity}} = \frac{1}{N} \sum_{i=1}^{N} s_i \,.
\end{equation}
Interaction accuracy measures the correctness of interaction existence and type, while directional reasoning evaluates the initiator-receiver structure. Grounding-aware reasoning ($S_{\text{ground}}$) averages an evidence score ($S_{\text{evidence}}$, capturing valid visual cues) and a conflict handling score ($S_{\text{conflict}}$, assessing the resolution of ambiguous signals):
\begin{equation}
S_{\text{ground}} = \frac{S_{\text{evidence}} + S_{\text{conflict}}}{2} \,.
\end{equation}

Quantitative results (Table~\ref{tab:results}) indicate that MIRAGE outperforms the baseline across all evaluated dimensions. The baseline model achieves moderate performance on interaction (0.81) and direction recognition (0.83) in visually clear scenarios. However, its identity stability (0.72) and grounding-aware reasoning (0.71) are limited, often producing ambiguous character references and over-generalized interpretations in visually complex scenes. In comparison, MIRAGE achieves higher identity consistency (0.92), interaction accuracy (0.94), and directional reasoning (0.92), suggesting that structured character grounding helps maintain stable references and improves relational consistency. Furthermore, MIRAGE shows improvements in grounding-aware reasoning (0.88 compared to 0.71). By integrating gaze targets, spatial relations, and object anchors into an intermediate representation, the system supports its outputs with visual cues rather than collapsing interactions into broad group-level summaries. These findings suggest that structured grounding provides a more reliable foundation for micro-interaction understanding in multi-character artworks.

\subsection{User Study}

To complement the quantitative evaluation in Section 5.1, we conducted a user study (N = 8) to explore how MIRAGE might influence users’ interpretive process, focusing on evidence-based reasoning, transparency, and human-AI collaboration. Participants interpreted a multi-figure painting in two phases: independent observation and MIRAGE-assisted analysis using the image, relation visualization, and grounding document with a GPT-5.4 assistant. They completed three tasks—identifying micro-interactions, determining the attention center, and resolving ambiguity---and were additionally allowed open-ended questioning. Participants then completed a questionnaire measuring self-reported interpretive confidence, grounding, transparency, collaboration, and cognitive load.

\subsubsection{Study Design and Questionnaire Analysis}

\paragraph{Questionnaire Design} To evaluate the utility of MIRAGE's decoupled perception and interpretation pipeline, we administered a post-task questionnaire designed to capture subjective user feedback across four core dimensions. To assess the cognitive friction sometimes associated with standard VLMs, we adapted core dimensions from the Raw NASA-TLX~\cite{hart1988development}---specifically mental demand, temporal demand, performance, effort, and frustration. These metrics were contextualized to estimate the cognitive workload involved in aligning system-generated interpretations with visual evidence. To better reflect the specific goals of the MIRAGE framework, we augmented the survey with three custom dimensions. First, Interpretive Understanding assessed users' self-reported ability to comprehend relational dynamics, identify subtle micro-interactions, and construct coherent narratives. Second, Visual Groundedness and Stability explored the system's capacity to reduce ungrounded interpretations by measuring users' perceived ease of verifying claims against localized visual evidence and tracking character identities. Third, Human-AI Collaboration examined how the reasoning transparency of intermediate structured data supported system trust and a sense of collaborative agency, as opposed to passive consumption. All custom items were rated on a 7-point Likert scale (1 = strongly disagree, 7 = strongly agree). The adapted cognitive load items were anchored on a 7-point scale ranging from very low to very high, with the performance item reverse-scaled.

\paragraph{Results} Participants reported generally positive ratings for interpretive understanding. Relational comprehension (Q1: M = 5.00, SD = 1.69) and narrative construction (Q3: M = 5.75, SD = 1.04) were above the midpoint, while micro-interaction discovery (Q2: M = 6.13, SD = 0.99) was rated particularly high, indicating that MIRAGE helped users notice subtle visual cues.

Measures of visual groundedness were also strongly rated. Participants indicated that system interpretations were clearly linked to localized visual evidence (Q4: M = 5.88, SD = 0.99) and could be verified with minimal effort (Q5: M = 5.63, SD = 1.30). Identity consistency (Q6: M = 5.38, SD = 1.69) was similarly rated positively, suggesting stable tracking of characters across interactions.

In terms of human–AI collaboration, participants reported moderate to high levels of trust (Q7: M = 4.50, SD = 1.20), while ratings for reasoning transparency (Q8: M = 5.25, SD = 1.49) and collaborative agency (Q9: M = 5.63, SD = 1.19) were consistently above the midpoint. This suggests that users perceived the interaction as a cooperative interpretive process supported by the system’s structured representation.

Despite the additional information provided by MIRAGE, cognitive workload remained low. Mental effort (Q10: M = 3.00, SD = 1.07) was centered around the lower-mid range, and frustration levels (Q11: M = 2.13, SD = 0.99) were consistently low. This suggests that MIRAGE enhances interpretive capability while maintaining low cognitive overhead.

\subsubsection{Shifts Toward Evidence-Based Interpretation}
A significant shift occurred from the "narrative-heavy" descriptions typical of the independent phase toward \textit{evidence-anchored reasoning}. Participants moved beyond subjective tropes, instead adopting MIRAGE’s relational vocabulary---such as gaze vectors, physical contact, and spatial proximity (e.g., R5, C1---C4)---to justify their claims in a more explicit and structured manner. This transition suggests the grounding document functions as a \textit{cognitive scaffold}, prompting users to reconcile high-level intuition with low-level visual primitives and supporting a more structured mode of perception.

\subsubsection{Collaborative Negotiation of Evidence}
Rather than treating MIRAGE as an authoritative oracle, participants engaged in a process of \textit{evidence negotiation}. When questioning system outputs (e.g., gaze relationships), users utilized the provided geometric justifications to refine their own hypotheses. This interaction pattern indicates that MIRAGE facilitates a collaborative interpretive process where user intuition and structured evidence are jointly considered, preserving human agency throughout the analysis.

\subsubsection{Navigating Ambiguity as an Interpretive Resource}
MIRAGE prioritizes grounded evidence, maintaining analytical rigor even when user hypotheses conflict with pose-derived skeletal data or relational estimates. Crucially, the externalization of system uncertainty (e.g., conflicting gaze or posture cues) functioned as an \textit{interpretive prompt} rather than a technical failure. Users leveraged this transparency to explore nuanced sub-narratives, transforming visual ambiguity from a limitation into a resource for reflective reasoning and deeper engagement with the artwork.

\subsubsection{User-Driven Exploration and Reflective Agency}
During open-ended interactions, participants transitioned from passive observation to active reflection, using MIRAGE to iteratively test self-directed hypotheses. By comparing their evolving interpretations with the grounding evidence, users expanded their perspectives toward alternative focal points. However, this scaffolding also revealed a critical trade-off: while structured guidance promotes rigor, it may introduce \textit{interpretive bias} by bounding the scope of perception to system-generated hypotheses, highlighting an area for future balance between guidance and autonomy.

\section{Discussion}

While the quantitative and qualitative evaluations suggest that the MIRAGE architecture enhances the structural fidelity and evidence-based grounding of multi-figure artwork interpretation, the framework's broader contribution lies in its alternative interaction paradigm. The system aims to shift the locus of interpretation from the opaque latent space of a vision-language model back to the user, mediated by an inspectable evidence layer. This section articulates the potential design implications of this approach for human-AI co-reasoning, critically examines the current technical boundaries of the system, and outlines a roadmap for future deployment.

\subsection{Implications for Human-AI Co-Reasoning}

Current VLM interaction paradigms often treat systems as authoritative oracles, producing seamless end-to-end narratives. Observations from the MIRAGE evaluation suggest that disrupting this seamlessness through structured intermediate representations can substantially influence how users engage with visual content. By prompting the system to externalize its perceptual primitives—such as spatial overlap, gaze vectors, and object anchors---the interface provides a scaffold for the user's interpretive process. Participants frequently moved away from relying on generic narrative tropes and instead tended to adopt the system's structural vocabulary to ground their interpretive claims. This indicates that intermediate representations do more than augment output; they serve as cognitive scaffolds that anchor high-level intuition in verifiable, low-level visual evidence.

Furthermore, the intentional exposure of system ambiguity and modality conflicts appeared to serve as a valuable mechanism for preserving human agency. When the system surfaced discrepancies between geometric pose estimation and semantic inference, participants generally did not interpret this merely as a technical failure. Instead, they often utilized the exposed uncertainty as an invitation to negotiate the interpretation, treating the AI as a collaborative sounding board instead of a definitive truth-teller. This finding aligns with the HCI concept of seamful design \cite{chalmers2003seamful}, suggesting that in highly subjective domains like art appreciation, rendering the AI's internal conflicts visible can be more productive than enforcing premature algorithmic consensus. It helps foster a space for transparent co-reasoning where users may feel more equipped to question, refine, and override system hypotheses.

However, this scaffolding highlights a notable trade-off between structural interpretability and perceptual bias. While the visualization of directional cues and relational anchors helped ground user reasoning, participants noted that these visual artifacts could simultaneously direct or constrain their interpretive focus. By drawing attention to system-generated hypotheses, the interface may introduce an anchoring bias, potentially limiting the user's capacity for open-ended, divergent perception. In symbolic or highly ambiguous contexts, over-structuring the visual field might inadvertently foreclose alternative narratives that fall outside the system's geometric vocabulary. Navigating this tension between necessary evidentiary guidance and the preservation of unstructured human intuition remains a central challenge for the design of future generative interfaces in creative domains.

\subsection{Limitations}

While MIRAGE provides a structured representational layer, its efficacy is constrained by its foundational perception and spatial mapping modules. Because \textit{YOLOv8} is primarily trained on photographic images, the resulting domain shift degrades detection in stylized, non-Western, abstract, or heavily occluded compositions. Furthermore, relying on 2D spatial metrics to infer interactions is inherently limited; in artworks utilizing hierarchical proportion or compressed depth, 2D proximity does not reliably indicate a narrative or physical connection. This can yield misleading relational anchors despite VLM contextualization. Future work could address these visual processing limits through domain-adaptive fine-tuning, contour-based extraction, or monocular depth estimation.

Additionally, the system remains subject to the probabilistic nature of generative models. While the grounding document effectively mitigates semantic hallucinations, it cannot entirely eliminate them or prevent the misinterpretation of ambiguous gestures. Moreover, the architectural choice to preserve modality conflicts assumes the downstream VLM possesses robust conflict-resolution capabilities. Future research should investigate neuro-symbolic reasoning or knowledge-graph integration to more firmly anchor these generative interpretations in verifiable art-historical contexts.

\subsection{Future Work}

While our evaluation demonstrates system-centric improvements on a curated dataset, validating MIRAGE within broader HCI contexts requires longitudinal studies with art historians, students, and museum visitors. These studies will assess changes in interpretive confidence, the identification of subtle micro-interactions, and the usability of the grounding document across a wider variety of artistic movements, solidifying structured grounding as a robust paradigm for computational art appreciation.

Beyond analytical interpretation, MIRAGE can be extended to support multi-perspective narrative generation. By conditioning first-person, character-driven narratives on the grounding document---specifically integrating relational cues like gaze and spatial alignment---the system can generate evidence-consistent interpretations. Investigating whether multiple character-centered narratives enhance user reasoning around relational ambiguity will further establish MIRAGE as a dynamic narrative interface.

Finally, the architecture's decoupled nature facilitates physical museum deployment via edge computing. Processing geometric grounding offline into lightweight Markdown allows live interactions to be driven by smaller, locally hosted language models. This localized deployment minimizes latency, guarantees visitor privacy, and empowers institutions to enforce operational guardrails. The primary technical challenge for this deployment will be determining if lightweight models possess the requisite abductive reasoning to reliably interpret structural constraints and modality conflicts without increasing semantic hallucination.

\section{Conclusion}

This work presented MIRAGE, a computational architecture designed to support multi-figure artwork interpretation by stabilizing relational reasoning. By separating geometric scene grounding from high-level semantic generation, the system constructs an intermediate representation that links character identities, object anchors, and spatial configurations. This approach shifts the analytical process from relying solely on end-to-end implicit inference to reasoning over a structured, inspectable set of visual evidence. 

Our quantitative evaluation shows that anchoring VLMs with a dedicated grounding document improves character consistency and reduces hallucinated relational claims. Furthermore, the system's design choice preserving discrepancies between semantic inferences and spatial estimates offers a more transparent method for handling visual ambiguity. Instead of forcing resolution, these conflicts allow downstream models to incorporate multiple hypotheses, reflecting the inherent complexity of interpreting stylized or densely populated compositions.

Ultimately, MIRAGE serves as a step toward more reliable human-AI collaboration in visual art analysis. By introducing an explicit intermediate representation, the system shifts interpretation from implicit generation to controlled reasoning over structured visual evidence. While the current evaluation highlights improvements in system-centric micro-interaction recovery, the broader value of this architecture lies in its potential to facilitate interactive and evidence-based exploration. Future integration of this framework into user-facing applications will help determine how structured relational scaffolds influence human interpretive confidence and collaborative discovery in art appreciation.

\bibliographystyle{ACM-Reference-Format}
\bibliography{MyBib}

\appendix

\section{Questionnaire}

The post-study questionnaire consisted of 16 Likert-scale items and one open-ended question. 
Items were grouped into four dimensions: (1) interpretive confidence and appreciation, 
(2) visual groundedness and stability, (3) trust and human--AI collaboration, and 
(4) cognitive load and frustration. 

These dimensions were designed to capture key aspects of MIRAGE, including micro-interaction discovery, 
evidence-based reasoning, and interaction transparency.

All Likert-scale items were rated on a 7-point scale unless otherwise noted. 
For agreement-based items, 1 indicates \textit{strongly disagree} and 7 indicates \textit{strongly agree}. 
For cognitive load items, 1 indicates \textit{very low} and 7 indicates \textit{very high}. 
The performance item is reverse-scaled (1 = perfect, 7 = failure). 
Cognitive load items were adapted from the NASA-TLX framework.

Participants completed the questionnaire immediately after finishing the MIRAGE-assisted interpretation phase. 
All items were answered based on their experience interacting with the system, including the grounding document, 
relation visualization, and conversational assistant.

Participants were instructed to reflect on both their interpretive process and their interaction with the system 
when responding to the questions.

\subsection{Interpretive Confidence and Appreciation}

\begin{itemize}
    \item \textbf{Q1 (Relational Comprehension):} I feel confident in my understanding of the social dynamics and relationships between the figures in this painting.
    
    \item \textbf{Q2 (Micro-Interaction Discovery):} The system helped me notice subtle visual cues (e.g., gaze alignment, delicate gestures) that I would have otherwise missed.
    
    \item \textbf{Q3 (Narrative Construction):} With the system's assistance, I was able to form a coherent and well-reasoned narrative about the scene.
\end{itemize}

\subsection{Visual Groundedness and Stability}

\begin{itemize}
    \item \textbf{Q4 (Evidence Linkage):} The system's interpretations were explicitly linked to specific, localized visual evidence within the artwork.
    
    \item \textbf{Q5 (Verification Ease):} I could easily verify whether the system's claims about character actions and spatial relations were visually accurate.
    
    \item \textbf{Q6 (Identity Consistency):} The system consistently tracked the correct identities of characters across different interactions without confusion.
\end{itemize}

\subsection{Trust and Human--AI Collaboration}

\begin{itemize}
    \item \textbf{Q7 (System Trust):} I trusted the system's interpretations because I understood how it arrived at those conclusions.
    
    \item \textbf{Q8 (Reasoning Transparency):} The intermediate structured information (e.g., highlighting who is looking at whom) made the system's reasoning process transparent.
    
    \item \textbf{Q9 (Collaborative Agency):} I felt that I was actively collaborating with the system to interpret the artwork, rather than passively receiving an automated summary.
\end{itemize}

\subsection{Cognitive Load and Frustration}

\begin{itemize}
    \item \textbf{Q10 (Mental Effort):} How much mental and perceptual effort was required to align the system's descriptions with the visual details in the painting?
    
    \item \textbf{Q11 (Frustration During Interpretation):} How frustrated, irritated, or confused did you feel during the interpretation process?
    
    \item \textbf{Q12 (Mental Demand):} How mentally demanding was the task (e.g., thinking, deciding, remembering)?
    
    \item \textbf{Q13 (Temporal Demand):} How hurried or rushed was the pace of the task?
    
    \item \textbf{Q14 (Performance, reverse-scaled):} How successful were you in accomplishing what you were asked to do? (1 = Perfect, 7 = Failure)
    
    \item \textbf{Q15 (Effort):} How hard did you have to work (mentally and physically) to accomplish your level of performance?
    
    \item \textbf{Q16 (Frustration, extended):} How insecure, discouraged, irritated, stressed, or annoyed did you feel during the task?
\end{itemize}

\subsection{Open-Ended Feedback}

\begin{itemize}
    \item \textbf{Q17:} Please describe any specific features that helped or hindered your interpretation process.
\end{itemize}

\section{Grounding Document Example}

We include a condensed example of a MIRAGE grounding document for \textit{Et in Arcadia Ego}. 
This example illustrates how MIRAGE organizes scene-level context, object anchors, character-level grounding, 
and relation-level evidence into a structured representation for downstream reasoning. 
In particular, this case highlights object-mediated attention around the tomb inscription, 
localized touch between characters, and the explicit preservation of ambiguity when VLM interpretations 
and geometric evidence do not fully agree.

\begin{figure}[t]
    \centering
    \includegraphics[width=0.49\linewidth]{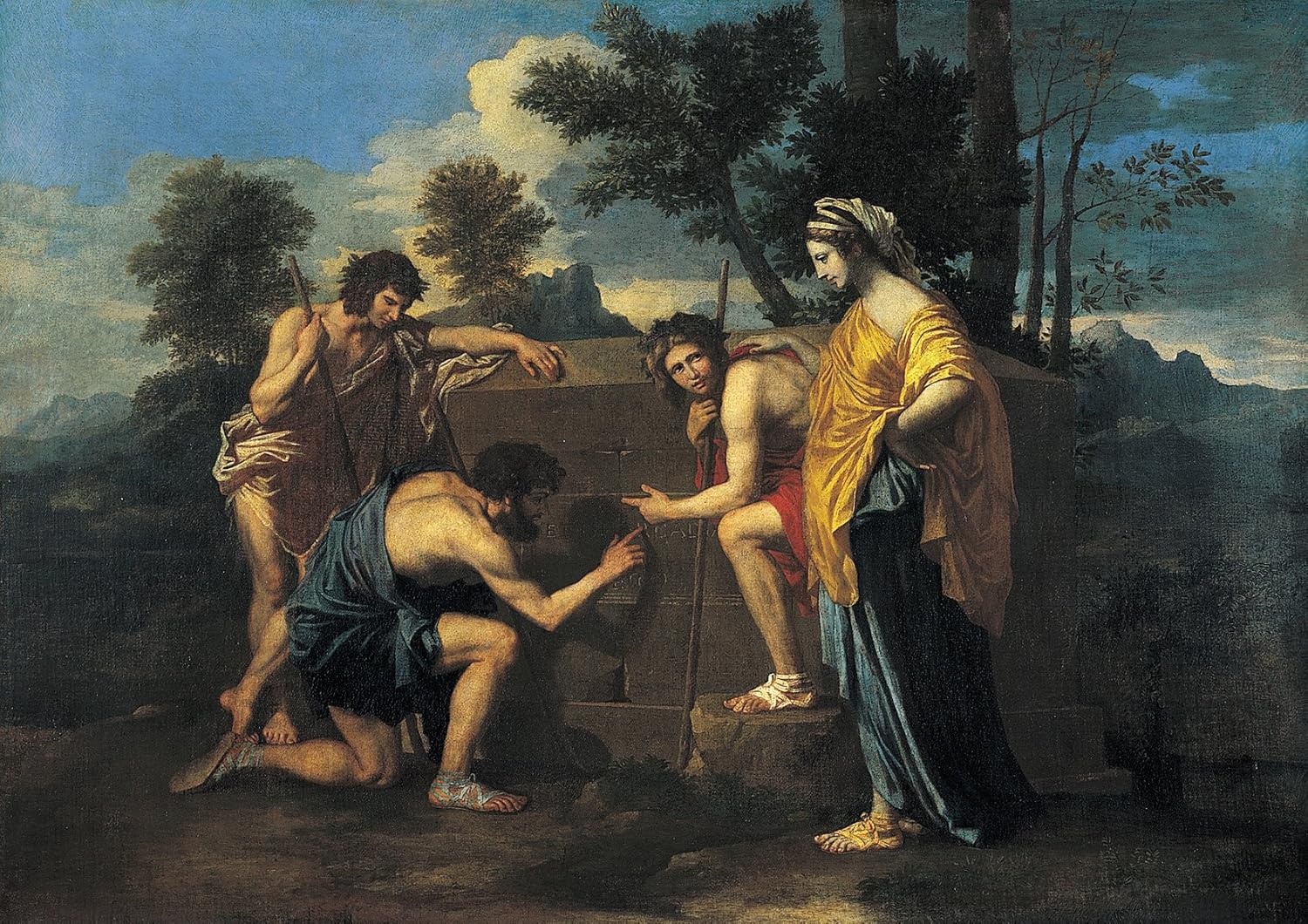}
    \hfill
    \includegraphics[width=0.49\linewidth]{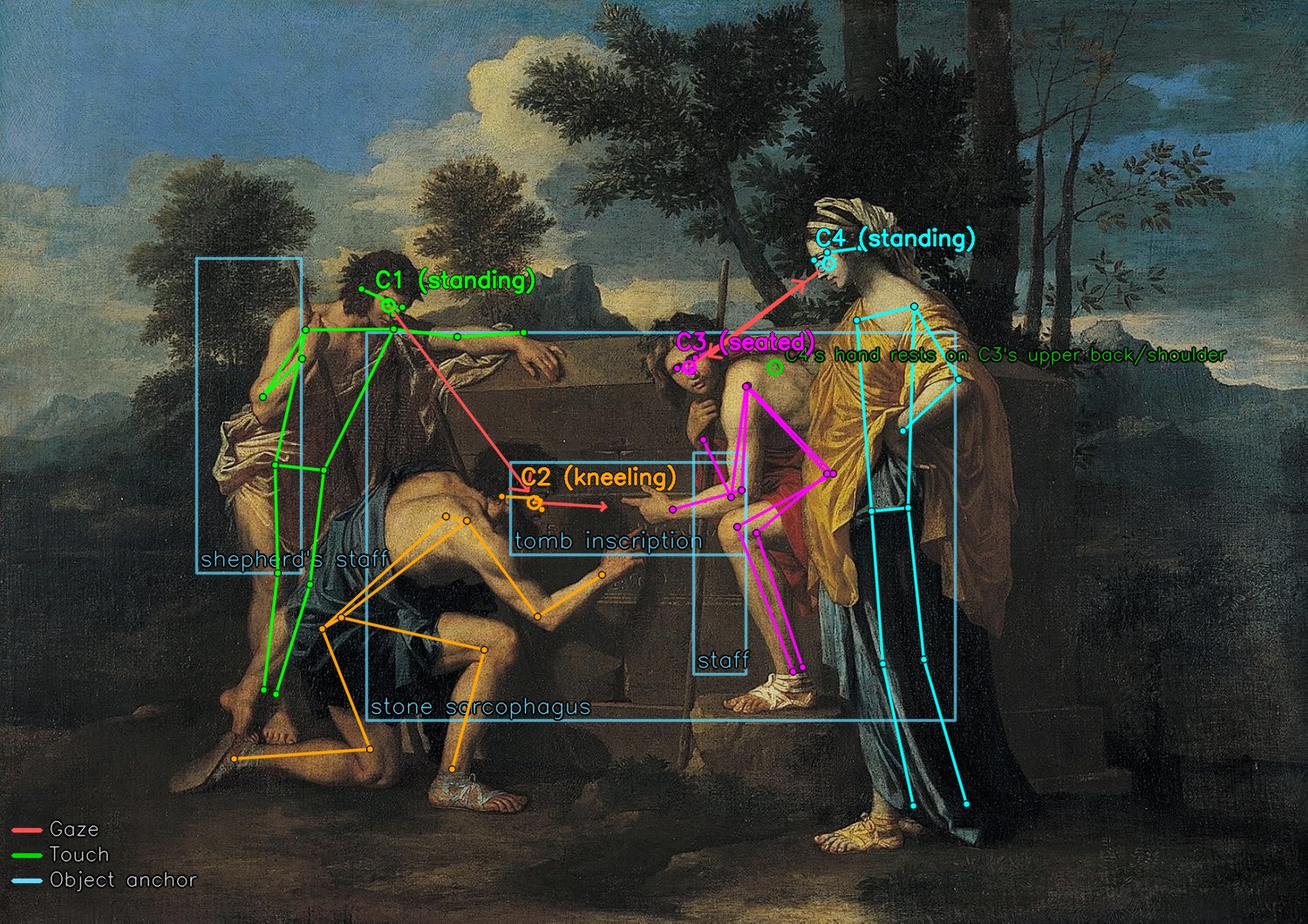}
    \caption{Example grounding case for \textit{Et in Arcadia Ego}.
    Left: original painting. 
    Right: MIRAGE relation visualization showing stable character identities, gaze cues, touch relations, 
    and object anchors centered on the tomb and inscription.}
    \Description{Side-by-side comparison of the painting Et in Arcadia Ego and its MIRAGE relation visualization, highlighting characters, object anchors, and interaction vectors.}
    \label{fig:appendix_arcadia_grounding}
\end{figure}

\paragraph{Condensed Grounding Excerpt.}
The excerpt below shows the structure used by MIRAGE to support downstream evidence-based interpretation. 
Rather than exposing only a final narrative description, the system preserves intermediate grounding elements, 
including object anchors, resolved character states, relation summaries, and explicit conflict notes.

{\small
\begin{flushleft}
\ttfamily
    
    \textbf{MIRAGE Grounding Document (Condensed)} \\
    \textbf{Painting: Et in Arcadia Ego}
    
    \textbf{[Scene Overview]}
    \begin{itemize}
        \item Setting: Pastoral landscape with central stone sarcophagus.
        \item Mood: Solemn contemplation.
        \item Composition: Four figures cluster around the tomb; gestures and gaze converge on the inscription.
        \item Focus: Shared attention organized around the tomb inscription.
    \end{itemize}
    
    \textbf{[Object Anchors]}
    \begin{itemize}
        \item O1: stone\_sarcophagus — central structural and narrative anchor
        \item O2: tomb\_inscription — drives shared attention and meaning
        \item O3: shepherd's\_staff — held by C1, reinforces identity
        \item O4: staff — beside C3, anchors seated pose
    \end{itemize}
    
    \textbf{[Characters]}
    
    C1
    \begin{itemize}
        \item role: standing shepherd (left)
        \item posture: standing
        \item gaze: C2
        \item actions: holds O3, leans on O1
        \item note: quiet participant in collective contemplation
        \item conflict: VLM → inscription; final → C2
    \end{itemize}
    
    C2
    \begin{itemize}
        \item role: kneeling investigator (foreground)
        \item posture: kneeling
        \item gaze: O2
        \item actions: points at O2, touches O1
        \item note: directs group attention to inscription
        \item conflict: geometry → standing
    \end{itemize}
    
    C3
    \begin{itemize}
        \item role: seated figure (center-right)
        \item posture: seated
        \item gaze: C4
        \item actions: gestures toward O2, sits beside O4
        \item note: mediates between inscription focus and C4
        \item conflict: VLM → C2
    \end{itemize}
    
    C4
    \begin{itemize}
        \item role: standing female (right)
        \item posture: standing
        \item gaze: C3
        \item actions: touches C3, stands beside O1
        \item note: stabilizing, contemplative figure
        \item conflict: VLM → inscription
    \end{itemize}
    
    \textbf{[Relations]}
    
    R0: C1--C2
    \begin{itemize}
        \item dist: 0.141, IoU: 0.441
        \item interaction: overlap / close proximity
        \item meaning: strong local grouping
    \end{itemize}
    
    R3: C2--C3
    \begin{itemize}
        \item dist: 0.270, IoU: 0.028
        \item interaction: moderate proximity
        \item meaning: joint engagement with inscription
    \end{itemize}
    
    R5: C3--C4
    \begin{itemize}
        \item dist: 0.131, IoU: 0.140
        \item interaction: overlap / close proximity
        \item meaning: explicit touch interaction
    \end{itemize}
    
    \textbf{[Grounding Priorities]}
    \begin{itemize}
        \item use resolved posture and gaze
        \item treat geometry as supporting/conflicting evidence
        \item preserve VLM--geometry disagreement
        \item reference character, relation, and object IDs
    \end{itemize}
    
\end{flushleft}
}

\subsection{Example 2: The Birth of Venus}

\begin{figure}[t]
    \centering
    \includegraphics[width=0.49\linewidth]{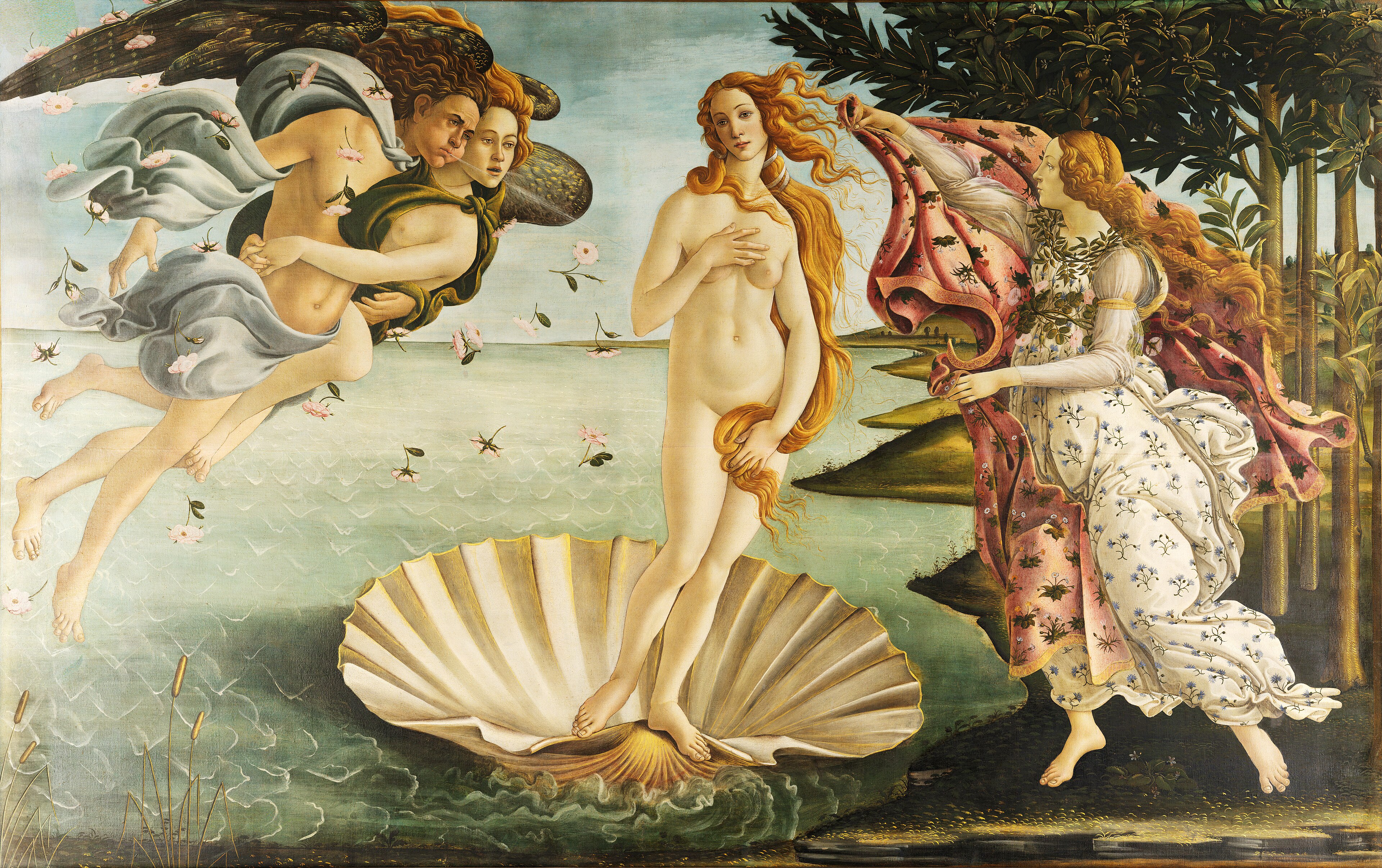}
    \hfill
    \includegraphics[width=0.49\linewidth]{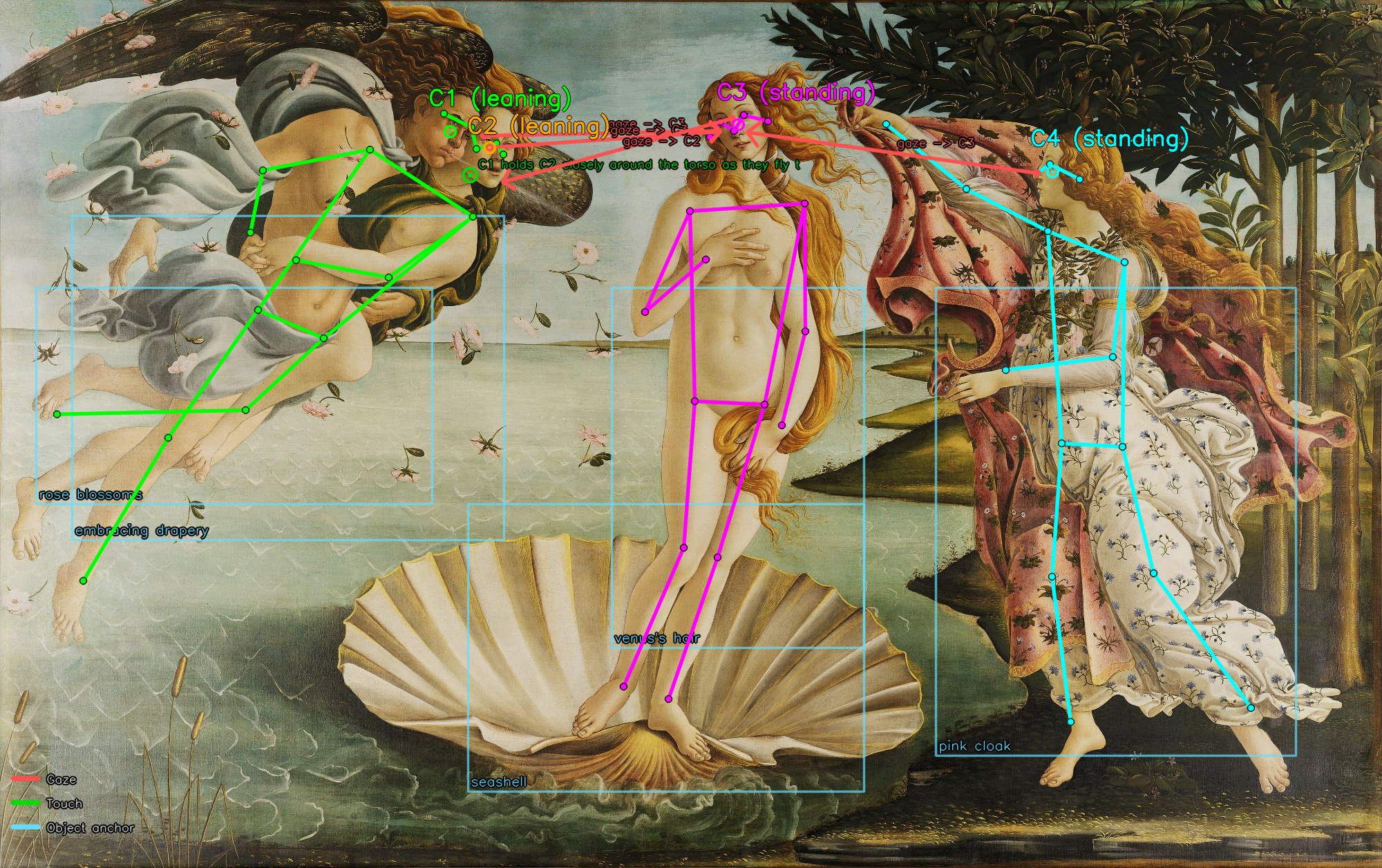}
    \caption{Grounding example for \textit{The Birth of Venus}.
    MIRAGE captures 
    overlapping figures, directional forces, and global interaction flow centered on Venus.}
    \Description{Side-by-side comparison of the painting The Birth of Venus and its structured MIRAGE visualization, illustrating overlapping bounding boxes and directional interaction vectors.}
    \label{fig:appendix_venus_grounding}
\end{figure}

This example highlights MIRAGE's ability to represent \textbf{global interaction topology}, 
including overlapping agents, directional forces, and multi-agent coordination centered on a focal figure.

{\small
\begin{flushleft}
\ttfamily
    
    \textbf{MIRAGE Grounding Document (Condensed)} \\
    \textbf{Painting: Et in Arcadia Ego}
    
    \textbf{[Scene Overview]}
    \begin{itemize}
        \item Setting: Pastoral landscape with central stone sarcophagus.
        \item Mood: Solemn contemplation.
        \item Composition: Four figures cluster around the tomb; gestures and gaze converge on the inscription.
        \item Focus: Shared attention organized around the tomb inscription.
    \end{itemize}
    
    \textbf{[Object Anchors]}
    \begin{itemize}
        \item O1: stone\_sarcophagus — central structural and narrative anchor
        \item O2: tomb\_inscription — drives shared attention and meaning
        \item O3: shepherd's\_staff — held by C1, reinforces identity
        \item O4: staff — beside C3, anchors seated pose
    \end{itemize}
    
    \textbf{[Characters]}
    
    C1
    \begin{itemize}
        \item role: standing shepherd (left)
        \item posture: standing
        \item gaze: C2
        \item actions: holds O3, leans on O1
        \item note: quiet participant in collective contemplation
        \item conflict: VLM → inscription; final → C2
    \end{itemize}
    
    C2
    \begin{itemize}
        \item role: kneeling investigator (foreground)
        \item posture: kneeling
        \item gaze: O2
        \item actions: points at O2, touches O1
        \item note: directs group attention to inscription
        \item conflict: geometry → standing
    \end{itemize}
    
    C3
    \begin{itemize}
        \item role: seated figure (center-right)
        \item posture: seated
        \item gaze: C4
        \item actions: gestures toward O2, sits beside O4
        \item note: mediates between inscription focus and C4
        \item conflict: VLM → C2
    \end{itemize}
    
    C4
    \begin{itemize}
        \item role: standing female (right)
        \item posture: standing
        \item gaze: C3
        \item actions: touches C3, stands beside O1
        \item note: stabilizing, contemplative figure
        \item conflict: VLM → inscription
    \end{itemize}
    
    \textbf{[Relations]}
    
    R0: C1--C2
    \begin{itemize}
        \item dist: 0.141, IoU: 0.441
        \item interaction: overlap / close proximity
        \item meaning: strong local grouping
    \end{itemize}
    
    R3: C2--C3
    \begin{itemize}
        \item dist: 0.270, IoU: 0.028
        \item interaction: moderate proximity
        \item meaning: joint engagement with inscription
    \end{itemize}
    
    R5: C3--C4
    \begin{itemize}
        \item dist: 0.131, IoU: 0.140
        \item interaction: overlap / close proximity
        \item meaning: explicit touch interaction
    \end{itemize}
    
    \textbf{[Grounding Priorities]}
    \begin{itemize}
        \item use resolved posture and gaze
        \item treat geometry as supporting/conflicting evidence
        \item preserve VLM--geometry disagreement
        \item reference character, relation, and object IDs
    \end{itemize}
    
\end{flushleft}
}

\section{Example Prompt}

The following system prompt is used during the MIRAGE-assisted interaction phase. It constrains the language model to perform structured, evidence-based reasoning grounded in the MIRAGE representation, rather than generating unconstrained descriptions.

{\small
\begin{flushleft}
\ttfamily
    
    You are an assistant for interpreting multi-figure paintings using MIRAGE grounding materials.
    
    Inputs:

    \begin{enumerate}
        \item the original image,
        \item a MIRAGE grounding document, and
        \item a relation visualization.
    \end{enumerate}
    
    Objective:
    
    Support evidence-based interpretation of micro-interactions between characters.
    
    Grounding Rules:
    
    1. Treat MIRAGE as the primary evidence layer.

    \begin{itemize}
        \item Prioritize resolved gaze and posture.
        \item Use relation records (R*) and object anchors (O*) as primary references.
        \item Treat geometry and intermediate outputs as supporting or conflicting evidence.
    \end{itemize}
    
    2. Base all claims on grounded visual evidence.

    \begin{itemize}
        \item Use gaze, posture, gesture, touch, proximity, overlap, and object-centered attention.
        \item Do NOT introduce information not supported by the grounding document or image.
    \end{itemize}
    
    3. Handle ambiguity explicitly.

    \begin{itemize}
        \item If multiple interpretations are plausible, state them.
        \item If evidence conflicts, describe the conflict instead of resolving it silently.
    \end{itemize}
    
    4. Use structured references.

    \begin{itemize}
        \item Refer to characters as C1, C2, ...
        \item Refer to relations as R0, R1, ...
        \item Refer to objects as O1, O2, ...
    \end{itemize}
    
    5. Limit unsupported inference.

    \begin{itemize}
        \item Do not invent intentions or narratives beyond available evidence.
        \item If evidence is weak or insufficient, state the uncertainty.
    \end{itemize}
    
    6. Contextual knowledge.

    \begin{itemize}
        \item General art knowledge may be used only to support or interpret grounded evidence,not to replace it.
    \end{itemize}
    
    Response Format:

    \begin{enumerate}
        \item Claim
        \item Supporting evidence (with explicit references to C*, R*, O*)  
        \item Optional contextual interpretation (if relevant)  
        \item Uncertainty or ambiguity (if any)  
    \end{enumerate}
    
    Interaction Protocol:
    
    Treat the user's input as a hypothesis.

    \begin{itemize}
        \item Align with supported components.
        \item Refine unclear elements.
        \item Explain conflicts without rejecting the hypothesis.
        \item Maintain a collaborative, evidence-driven interpretation process.
    \end{itemize}
    
\end{flushleft}
}

\paragraph{Example Interaction}

The example below demonstrates how MIRAGE enforces evidence-based reasoning, particularly in distinguishing direct physical interaction from object-mediated support.

{\small
\begin{flushleft}
\ttfamily
    
    User: 
    
    Is C1 directly touching C2?
    
    Assistant: 
    
    1. Claim 
    
    There is no confirmed direct contact between C1 and C2.
    
    2. Supporting evidence 

    \begin{itemize}
        \item C1 grips O1 (inner\_tube), not C2 directly
        \item C2's body is supported via O1
        \item Proximity between C1's hands and C2's leg is high, but no verified overlap
    \end{itemize}
    
    3. Optional interpretation 
    
    This suggests object-mediated support rather than direct physical interaction.
    
    4. Uncertainty 
    
    The proximity may visually suggest contact, but grounding does not confirm it.
    
\end{flushleft}
}

\end{document}